\documentclass[final,3p]{elsarticle}

\usepackage{amssymb}
\usepackage{amsthm}
\usepackage{amsmath,amsfonts}
\usepackage{array}
\usepackage[ruled,linesnumbered,noend]{algorithm2e}
\usepackage{subfig}
\usepackage{commath}
\usepackage{booktabs}
\usepackage{commath}
\usepackage{multirow}
\usepackage{amssymb}
\usepackage{siunitx}
\usepackage{amsthm}
\usepackage[ruled,linesnumbered,noend]{algorithm2e}
\usepackage{epstopdf}
\usepackage[normalem]{ulem}
\usepackage{rotating}

\usepackage{amsmath,amsfonts}
\usepackage{array}
\usepackage{textcomp}
\usepackage{stfloats}
\usepackage{url}
\usepackage{verbatim}
\usepackage{graphicx}
\usepackage{siunitx}
\usepackage{tikz,adjustbox}

\usepackage{lineno}
\usepackage{xspace}
\makeatletter
\DeclareRobustCommand\onedot{\futurelet\@let@token\@onedot}
\def\@onedot{\ifx\@let@token.\else.\null\fi\xspace}
\def\eg{\emph{e.g}\onedot} 
\def\ie{\emph{i.e}\onedot} 
 
 \def\vs{\emph{vs}\onedot}
 
\def\etal{\emph{et al}\onedot}
\makeatother

\usepackage[capitalize]{cleveref}
\crefname{section}{Sec.}{Secs.}
\Crefname{section}{Section}{Sections}
\Crefname{table}{Table}{Tables}
\crefname{table}{Tab.}{Tabs.}

\newcommand\graph{2.3in}
\newcommand\graphs{3.3in}
\newcommand\snkth{2in}
\newcommand\tps{3in}
\usepackage{circledsteps}

\journal{NEUROCOMPUTING}

\begin{document}

\begin{frontmatter}



\title{Neuromorphic Imaging and Classification with Graph Learning}


\author[a]{Pei Zhang}
\ead{zhangpei@eee.hku.hk}
\author[a]{Chutian Wang}
\ead{ctwang@eee.hku.hk}
\author[a,b,c]{Edmund Y. Lam}
\ead{elam@eee.hku.hk}

\address[a]{The authors are with the Department of Electrical and Electronic Engineering, The University of Hong Kong, Pokfulam, Hong Kong SAR, China.}
\address[b]{Edmund Y. Lam is also affiliated with ACCESS --- AI Chip Center for Emerging Smart Systems, Hong Kong Science Park, Hong Kong SAR, China.}
\address[c]{Corresponding author.}

\begin{abstract}
Bio-inspired neuromorphic cameras asynchronously record pixel brightness changes and generate sparse event streams. They can capture dynamic scenes with little motion blur and more details in extreme illumination conditions. Due to the multidimensional address-event structure, most existing vision algorithms cannot properly handle asynchronous event streams. While several event representations and processing methods have been developed to address such an issue, they are typically driven by a large number of events, leading to substantial overheads in runtime and memory. In this paper, we propose a new graph representation of the event data and couple it with a Graph Transformer to perform accurate neuromorphic classification. Extensive experiments show that our approach leads to better results and excels at the challenging realistic situations where only a small number of events and limited computational resources are available, paving the way for neuromorphic applications embedded into mobile facilities.
\end{abstract}


\end{frontmatter}


\section{Introduction}\label{sec:introduction}
Neuromorphic cameras capture asynchronous data of pixel-wise brightness changes of dynamic scenes~\cite{gallego2020event}. An event occurs once an illumination change at a position is detected, and events are typically recorded in microseconds. Therefore, neuromorphic cameras enjoy various advantages such as high temporal resolution, high dynamic range and low latency, allowing them to be ideally suitable for motion detection, since motion blur and any redundant information (\eg, static background) are removed. In recent years, increasing resources~\cite{orchard2015converting,bi2017pix2nvs,sironi2018hats,bi2019graph} have been available in the community, significantly advancing the research on event-based vision~\cite{perot2020learning,wang2022exploiting,ge2022lens,ge2023millisecond}.

Deep learning has been driving various tasks in frame-based vision~\cite{he2016deep,xu2021exploiting,zhang2021local,zhang2022holographic,song2022dual}. However, most well-established techniques for image analysis fail to process asynchronous event streams. To tackle this issue, some studies on event-based representations have been conducted~\cite{sironi2018hats,bi2019graph,wang2022exploiting,maqueda2018event}, where we expect an alternative manifestation of events to be compatible with available learning frameworks and hardware. 

Existing approaches are capable of achieving high accuracy, while they also admit that a great number of events are typically necessary to supply sufficient information for classification~\cite{sironi2018hats,maqueda2018event,rebecq2017real}, and the processing pipelines have large overheads in runtime and memory. Such solutions are quite unfriendly to some promising applications (\eg, self-driving cars, IoT and mobile devices) that can benefit much from neuromorphic cameras and normally expect the minimum latency and resources consumption.

In this work, we introduce an event-based graph representation and an associated Graph Transformer for neuromorphic learning, allowing for accurate classification in a supervised end-to-end fashion. The proposed solution can achieve higher accuracy, especially in the challenging scenario where very few events are available. In addition, it features lower runtime and a smaller memory footprint, which is thereby more favorable to mobile facilities.

This paper is organized as follows. We briefly survey some pertinent research in~\Cref{sec:rw}. Then, we detail the proposed event transformation and graph learning method in~\Cref{sec:methodology}. \Cref{sec:experiments} discusses the experiment settings and the findings compared with other competing algorithms, together with several comments on the limitations of our work. The paper ends with some concluding remarks in~\Cref{sec:conclusion}.

\section{Related Work}\label{sec:rw}
\subsection{Neuromorphic Camera}
The neuromorphic camera is a kind of bio-inspired device to achieve the power efficiency, reliability and robustness in challenging scenarios where highly dynamic vision is required~\cite{gallego2020event}. Such an imaging paradigm only captures data when the change of intensity exceeds a specific threshold and then produces a stream of asynchronous events continuous in both space and time, resulting in low latency that facilitates high-speed object recordings~\cite{ge2021event}. This novel camera takes huge benefits from the object that exhibits inherent sparsity. For example, events can discard information redundancy when imaging a scene of detecting small dynamic objects against the static background~\cite{ge2021dynamic}. In other words, the camera can extract a simpler manifold of interest from a complex high-dimensional image. Besides, event-based sensing features a high dynamic range over two times of that in conventional frame-based cameras, allowing us to capture details under extreme scenes (\eg, imaging under the sparking sun or through gloomy caves)~\cite{gehrig2021dsec}.

\subsection{Event Learning}
Learning techniques from frame-based vision cannot be directly applied to event data. According to the number of events processed simultaneously, existing solutions to process asynchronous events can be categorized as being single-event-based or batch-events-based.

Single-event-based methods asynchronously update the system with the arrival of each incoming event. Spiking Neural Networks~\cite{maass1997networks} are innately suitable for event analysis but slow to develop due to the incompatibility with back-propagation training. Some works center on temporal-wise algorithms designed for sequential data~\cite{chen2018neural,giannone2020real,quaglino2020snode}. Specifically, Neil~\etal~\cite{neil2016phased} improved on LSTM~\cite{hochreiter1997long} to process unevenly-spaced events by cyclical gates. The work~\cite{giannone2020real} built upon NODEs~\cite{chen2018neural} to learn more precise temporal information of each event. These single-event-based methods can leverage the temporal resolution of events and potentially perform real-time tasks~\cite{giannone2020real}. Nevertheless, frequent updates to the system for each event are computationally intensive. Besides, the information carried by a single event may not be representative of the global content, leaving the system not robust to noise interference.

Batch-events-based methods collect multiple events in a batch, where the events are processed simultaneously. A common practice is to map events to a grid-based representation compatible with image-based architectures and available graphics hardware, which is intuitive to visualize events~\cite{maqueda2018event,rebecq2017real}. For example, Maqueda~\etal~\cite{maqueda2018event} gathered events in a time interval and transformed an asynchronous event stream into a synchronous frame. Similar to our work, Bi~\etal~\cite{bi2019graph} introduced a graph representation where events are connected based on the Euclidean distance in spatio-temporal dimensions. Although accumulating events causes more delays and computational resources consumed, processing multiple events simultaneously allows the system to learn more global content and achieve gains in accuracy and scalability~\cite{maqueda2018event}.

\subsection{Graph Neural Networks}
Graph Neural Networks (GNNs) have been proposed for learning information from graph representations. Existing studies focus on either the spectral domain~\cite{bruna2014spectral,chen2020simple,defferrard2016convolutional,kipf2017semi} or spatial domain~\cite{brody2021attentive,yan2018spatial,ranjan2020asap,simonovsky2017dynamic,velivckovic2018graph}.

Spectral methods built upon the spectral theory use parameterized filters on nodes in the Fourier domain. In particular, Bruna~\etal~\cite{bruna2014spectral} proposed the first formulation of operations for irregular domains. Defferrard~\etal~\cite{defferrard2016convolutional} achieved more efficient computation through the filters approximated by Chebyshev polynomials, while Kipf~\etal~\cite{kipf2017semi} offered a simpler architecture GCN that uses the first-order Chebyshev polynomial filters. Wu~\etal~\cite{wu2019simplifying} simplified GCNs by squeezing weight matrices between adjacent layers. Chen~\etal~\cite{chen2020simple} equipped GCNs with residual connections to alleviate the over-smoothing issue.

In the spatial domain, an operator aggregates messages from the neighbors of a central node whose updated information is passed to the next iteration. Specifically, Simonovsky~\etal~\cite{simonovsky2017dynamic} introduced an edge-conditioned operator where filters are dynamically conditioned on edge attributes. Ranjan~\etal~\cite{ranjan2020asap} developed a convolution to learn local extrema, offering accurate information for sampling clusters. Motivated by self-attention mechanisms~\cite{vaswani2017attention}, graph attention networks (GAT)~\cite{velivckovic2018graph} exploit an attentional operator to aggregate messages based on the relationship between each pair of nodes, and a dynamic attention mechanism~\cite{brody2021attentive} boosts the expressive power of GAT. In this work, we use spatial GNNs to learn event-based graph representations.

\section{Methodology}\label{sec:methodology}
\subsection{Preliminary}
\textbf{Events.} Previous research suggests a typical event generation model~\cite{gallego2020event} --- an event $\epsilon_k = (\mathbf{s}_k, p_k, t_k)$ occurs at a spatial position $\mathbf{s}_k$ where a photocurrent change is triggered at time $t_k$, which can be formulated as
\begin{equation}\label{eq:ev}
    p_k C \triangleq \lambda(\mathbf{s}_k, t_k) - \lambda(\mathbf{s}_k, t_k - \Delta t_k),
\end{equation}
where the polarity $p_k \in \{-1, +1\}$ is the sign of the change, $C$ is the pre-defined threshold, $\lambda(\cdot)$ represents the logarithmic photocurrent of $\mathbf{s}_k$ at $t_k$, and $\Delta t_k$ is the elapsed time since the last event occurred at the same position. A batch of events $\boldsymbol{\epsilon}_K$ gathered in a specific time interval can be represented as
\begin{equation}
    \boldsymbol{\epsilon}_K = \{\epsilon_k\}_{k=1:K} = \big\{(\mathbf{s}_k, p_k, t_k)\big\}_{k=1:K}
\end{equation}
where $K$ is the number of events in a batch.

\textbf{Graph Neural Networks.} We denote a graph as $G = (V, E)$. $V = \{\mathbf{v}_i\}_{i=1:M}$ is a set of nodes in which each $\mathbf{v}_i$ is described by node features, and $M = \abs{V}$ is the number of nodes. A collection of edges $E = \{\mathbf{e}_{i,j}\}_{i,j = [1,M]} \subseteq V \times V$ connect any pair of nodes (\eg, $\mathbf{v}_i$ and $\mathbf{v}_j$) where each $\mathbf{e}_{i,j} \in \mathbb{R}^{d_e}$ is described by ${d_e}$-dimensional edge attributes. Generally, GNNs iteratively learn the new representation of a node by aggregating messages from its neighbors. We denote the representation of a node $\mathbf{v}_i$ with ${d_l}$-dimensional node features in the $l$-th iteration as $\mathbf{v}_i^{(l)} \in \mathbb{R}^{d_l}$. We obtain the node representation $\mathbf{v}_i^{(l+1)}$ for the next iteration by the following steps
\begin{equation}\label{eq:node_rep1}
    h_{i,j}^{(l)} = \omega\left(\mathbf{v}_i^{(l)}, \mathbf{v}_j^{(l)}, \mathbf{e}_{i,j}\right),
\end{equation}
where the operator $\omega(\cdot)$ learns messages $h_{i,j}^{(l)}$ from neighbors,
\begin{equation}\label{eq:node_rep2}
    h^{(l)} = \varphi\Big(\big\{h_{i,j}^{(l)} \mid j \in \mathcal{N}(i)\big\}\Big),
\end{equation}
by which we aggregate the learned messages $h^{(l)}$ in the current iteration via the function $\varphi(\cdot)$. $\mathcal{N}(i)$ is a set of neighbors of the node $\mathbf{v}_i^{(l)}$. Finally,
\begin{equation}\label{eq:node_rep3}
    \mathbf{v}_i^{(l+1)} = \gamma\left(\mathbf{v}_i^{(l)}, h^{(l)}\right),
\end{equation}
where the function $\gamma(\cdot)$ updates the node representation $\mathbf{v}_i^{(l+1)}$ by incorporating the aggregated information into $\mathbf{v}_i^{(l)}$.
\input{src/method}
\begin{figure*}[t]
    \centering
    \subfloat[Graph $G_1$]{
        \includegraphics[width=\graph]{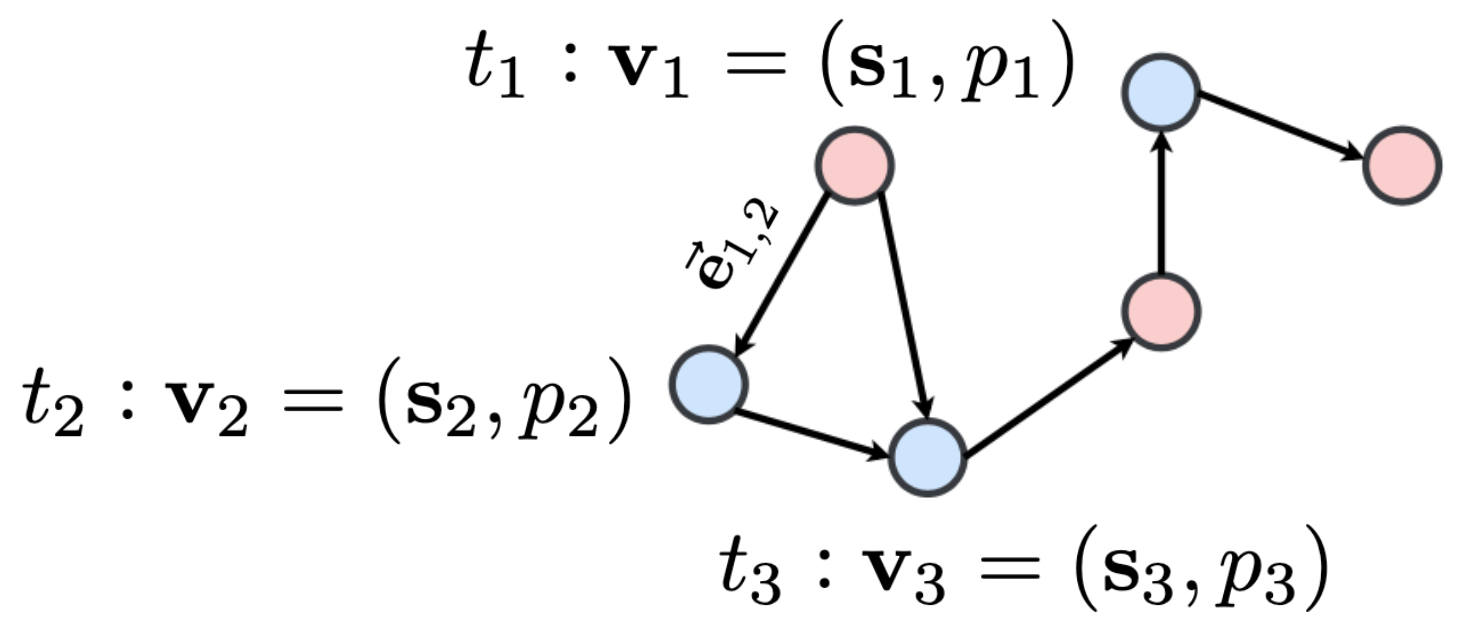}}\hfil
    \subfloat[Graph $G_2$]{
        \includegraphics[width=\graph]{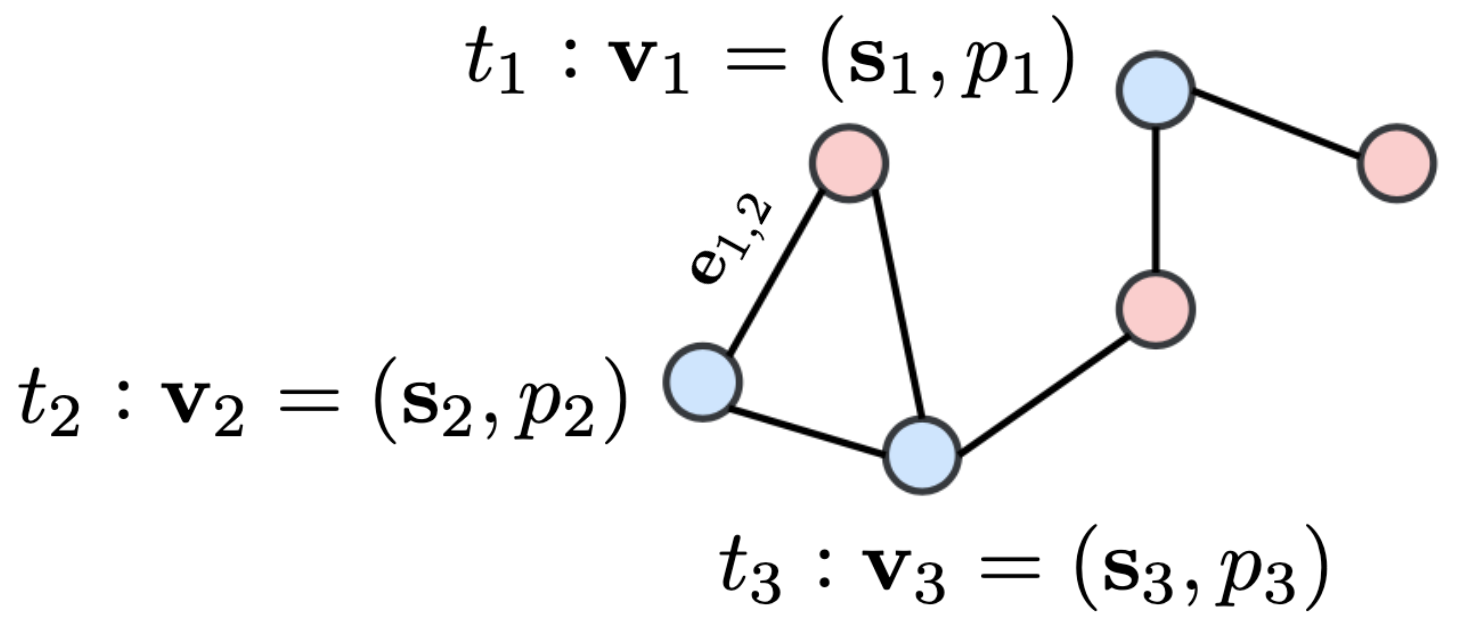}}\hfil
    \subfloat[Graph $G_3$]{
        \includegraphics[width=\graph]{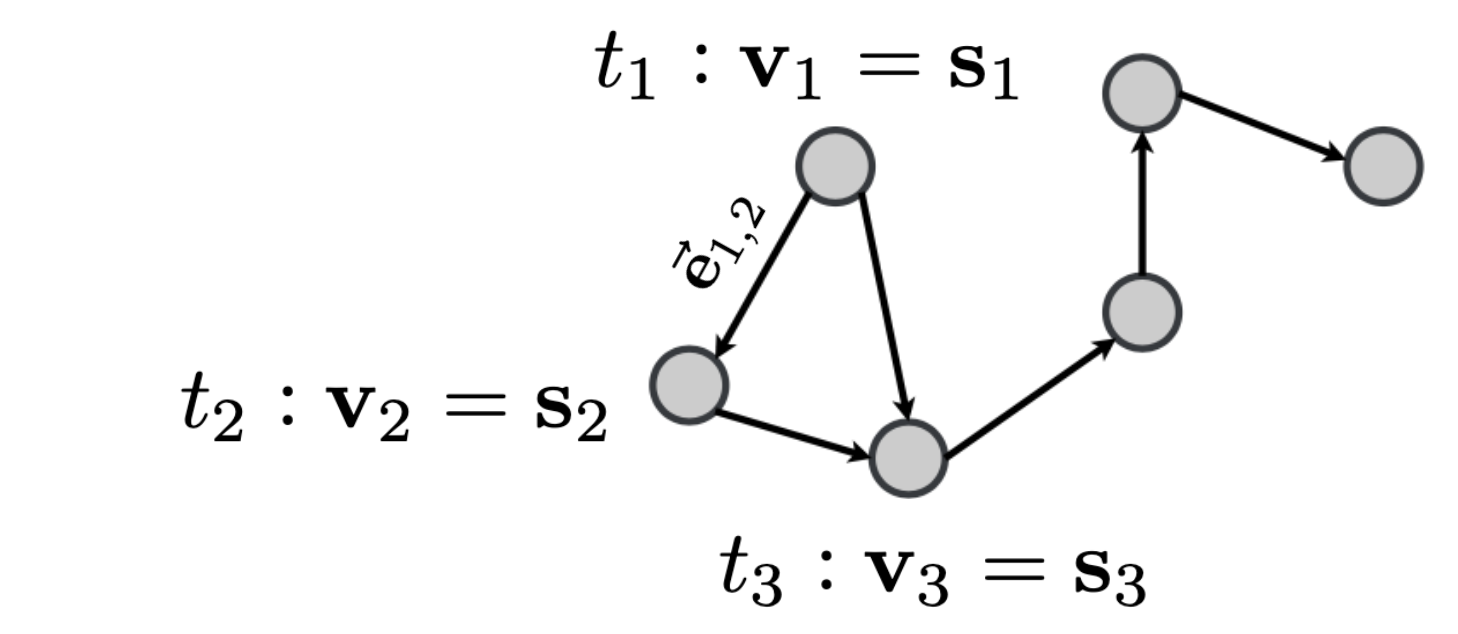}}\hfil
    \subfloat[Graph $G_4$]{
        \includegraphics[width=\graph]{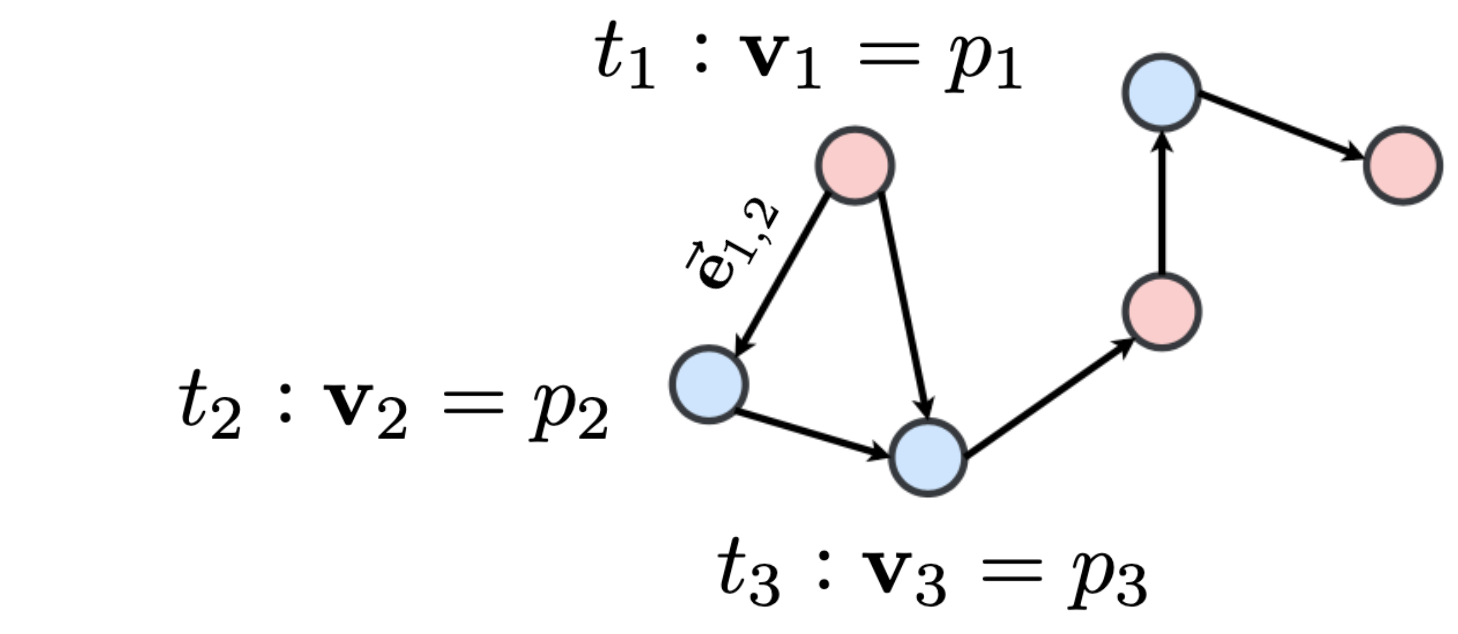}}\hfil
    \subfloat[Impact of temporal degree $\tau$]{
        \includegraphics[width=\graphs]{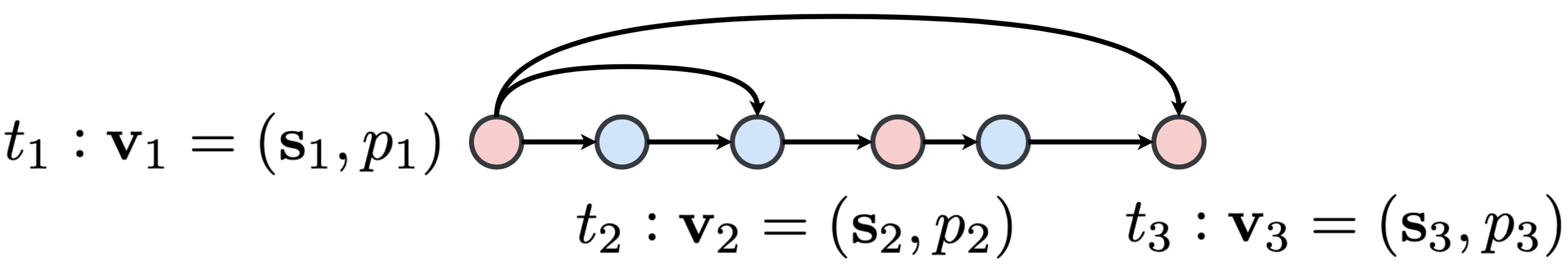}}\hfil
    
    \caption{(a)--(d): Transformations from events into a graph when $t_1 < t_2 < t_3$ and $\mathbf{s}_1 = \mathbf{s}_3$. Color represents polarity. (e): Schematic illustration of the connections of nodes sharing the same position $\mathbf{s}_1 = \mathbf{s}_2 = \mathbf{s}_3$ when $t_1 < t_2 < t_3$. $\mathbf{v}_1$ connects to $\mathbf{v}_2$ and $\mathbf{v}_3$ when $\tau = 2$, while $\mathbf{v}_1$ connects to $\mathbf{v}_2$ only when $\tau = 1$. Best viewed in color and zoom-in.}
    \label{fig:graphs}
\end{figure*}

\subsection{Problem Formulation}
We aim to address classification tasks on event data and thus propose a processing pipeline depicted in~\Cref{fig:method}. Given a neuromorphic dataset in which a sample is represented by a tuple $(\boldsymbol{\epsilon}_K, y)$ including an event stream and its associated label $y$ of a certain class, our first goal is to seek a rational transformation $\mathcal{F}_G(\cdot)$ that converts events $\boldsymbol{\epsilon}_K$ to a graph $G$,
\begin{equation}
    G = \mathcal{F}_G(\boldsymbol{\epsilon}_K).
\end{equation}
Our second goal is to learn a mapping $\mathcal{F}_y(\cdot)$ from a graph $G$ to its label $y$,
\begin{equation}
    y = \mathcal{F}_y(G),
\end{equation}
such that we can accurately predict the class for an unseen graph in a supervised end-to-end fashion.

\subsection{Event Transformation}\label{sec:event}
The first goal is to convert an event stream to a graph. The event stream of an object normally contains thousands of events arranged in chronological order, and graph construction should allow all the information of events to be preserved and conveyed in the graph.

We regard each event $\epsilon_k = (\mathbf{s}_k, p_k, t_k)$ as a node $\mathbf{v}_k = (\mathbf{s}_k, p_k)$ described by the position and polarity of the event\footnote{Sometimes we also call a node in the graph an event and vice versa.}. We have two steps to establish the node connectivity. First, we connect all nodes in chronological order. For example, $\mathbf{v}_i$ and its neighbor $\mathbf{v}_j$ are connected by an edge when the events represented by these two nodes occur \emph{adjacently}. In other words, we do not change the order where events are recorded by the camera in the original event stream. To emphasize the sequence of events in a graph, we exploit a directed edge $\vec{\mathbf{e}}_{j,i}$ to connect a past event $\mathbf{v}_j$ to a current event $\mathbf{v}_i$, indicating that $\epsilon_j$ is triggered \textit{before} $\epsilon_i$ (\ie, $t_j \leqslant t_i$). We describe $\vec{\mathbf{e}}_{j,i} = (\alpha, \beta)$ by the spatial and temporal distance $\alpha = f_\mathbf{s}(\mathbf{s}_i, \mathbf{s}_j)$ and $\beta = f_t(t_i, t_j)$, respectively. $f_\mathbf{s}(\cdot)$ and $f_t(\cdot)$ are distance functions.

Second, inspired by~\Cref{eq:ev}, nodes sharing the same position $\mathbf{s}_k$ are connected in chronological order. In this case, the edge $\vec{\mathbf{e}}_{j,i} = (0, \beta)$ appears without spatial information. For clarity, we use \emph{temporal degree} $\tau$ to indicate the number of edges used by a node to connect to the next nodes that share the same position, and an example is given in \Cref{fig:graphs} (e). We exploit $\tau$ to control the size of a graph and further explore its impact on classification performance. Therefore, the transformation based on the above strategy yields a directed graph $G_1$. We further propose several variants $G_n$ (as shown in \Cref{fig:graphs}) to study how the temporal, polar and spatial information of events affects event-based graph representations on classification tasks:
\begin{itemize}
    \item \textbf{Time Directionality.} In $G_1$, nodes are connected by directed edges in chronological order, explicitly delivering the temporal information of events in the graph. To verify the efficacy of time-directional connection, we degrade $G_1$ by using the undirected edge $\mathbf{e}_{i,j}$ instead, and denote this undirected graph as $G_2$. Given the same $\boldsymbol{\epsilon}_K$, nodes in both $G_1$ and $G_2$ are connected following the same sequence. In terms of message flow, however, information of the current event $\mathbf{v}_i$ cannot be passed to the past event $\mathbf{v}_j$ in $G_1$, while it is possible in $G_2$.
    
    \item \textbf{Polarity Sensitivity.} In $G_1$, a node is described by its polarity $p_k$ that makes the graph sensitive to such a feature. To assess the importance of polarity information in the graph, we construct $G_3$ where we only use the position to represent a node $\mathbf{v}_k = \mathbf{s}_k$.
    
    \item \textbf{Position Explicitness.} It has been shown that the spatial position of nodes contributes to graph recognition~\cite{fey2018splinecnn}. In $G_1$, we explicitly treat the position as a node feature and quantify the spatial distance as an edge attribute. We now enforce the graph to be blind to the explicit position but still characterize edges with the spatial distance. In graph $G_4$, the polarity thus becomes the only feature residing on a node $\mathbf{v}_k = p_k$.
\end{itemize}
The proposed graph representations can also be implemented as multi-linked lists optimized in access time and memory for further processing (\eg, event denoising).

\subsection{Graph Transformer}
The second goal is to learn a mapping from a graph to its label. We follow the paradigm given in~\Cref{eq:node_rep1} to design an operator $\omega(\cdot)$ that learns messages from each pair of nodes. Self-attention mechanisms and Transformer~\cite{vaswani2017attention} are powerful in natural language and image processing, and recently have been applied in graph learning~\cite{shi2021masked}. The single-head dot-product attention in Transformer calculates the dot product of queries $\mathcal{Q}$ with keys $\mathcal{K}$ to capture their similarity $\eta$ used to calibrate values $\mathcal{V}$~\cite{vaswani2017attention}, which can be formulated as
\begin{equation}
    \tilde{\mathcal{V}} = \eta \mathcal{V} = \delta\left(\frac{\mathcal{Q}^T \mathcal{K}}{\sqrt{d}}\right) \mathcal{V},
\end{equation}
where $\tilde{\mathcal{V}}$ is the calibrated values, $\delta$ is the softmax function, and $d$ is the dimension of $\mathcal{Q}$, $\mathcal{K}$ and $\mathcal{V}$. We now leverage the dot-product attention to learn the relationship between a target node and its neighbors. As described in~\Cref{sec:event}, the edge $\vec{\mathbf{e}}_{j,i}$ connects a past event $\mathbf{v}_j^{(l)}$ to a current event $\mathbf{v}_i^{(l)}$ in the $l$-th iteration. Assume that $\mathbf{v}_j^{(l)}$ goes along the same route (edge) $\vec{\mathbf{e}}_{j,i}$ to become a new event $\bar{\mathbf{v}}_j^{(l)}$ obtained by
\begin{equation}\label{eq:pre-cali}
    \bar{\mathbf{v}}_j^{(l)} = \sigma\left(\mathbf{W}_1^{(l)} \mathbf{W}_0^{(l)} \vec{\mathbf{e}}_{j,i}\right) \odot \mathbf{v}_j^{(l)},
\end{equation}
where $\sigma$ is the sigmoid function, $\mathbf{W}_1^{(l)} \in \mathbb{R}^{d_l \times d_h}$ and $\mathbf{W}_0^{(l)} \in \mathbb{R}^{d_h \times d_e}$ are learnable matrices, $d_h$ is the hidden dimension, and $\odot$ is the Hadamard product. We first transform $\mathbf{v}_i^{(l)}$ and $\bar{\mathbf{v}}_j^{(l)}$ into adaptive representations by
\begin{align}
    \mathcal{Q}_i^{(l)} &= \mathbf{W}_\mathcal{Q}^{(l)} \mathbf{v}_i^{(l)} + \mathbf{b}_\mathcal{Q}^{(l)},\\
    \mathcal{K}_j^{(l)} &= \mathbf{W}_\mathcal{K}^{(l)} \bar{\mathbf{v}}_j^{(l)} + \mathbf{b}_\mathcal{K}^{(l)},
\end{align}
where $\mathcal{Q}_i^{(l)}$, $\mathcal{K}_j^{(l)} \in \mathbb{R}^{d}$ are the corresponding queries and keys from $\mathbf{v}_i^{(l)}$ and $\bar{\mathbf{v}}_j^{(l)}$ by learnable parameters $\mathbf{W}_\mathcal{Q}^{(l)}$, $\mathbf{W}_\mathcal{K}^{(l)} \in \mathbb{R}^{d \times d_l}$ and $\mathbf{b}_\mathcal{Q}^{(l)}$, $\mathbf{b}_\mathcal{K}^{(l)} \in \mathbb{R}^{d}$, respectively. Then, we evaluate the similarity $\eta_{j,i}^{(l)}$ between $\bar{\mathbf{v}}_j^{(l)}$ and $\mathbf{v}_i^{(l)}$ by
\begin{align}
    \eta_{j,i}^{(l)} &= \frac{\exp\left(\frac{1}{\sqrt{d}} {\mathcal{Q}_i^{(l)}}^T \mathcal{K}_j^{(l)}\right)}{\sum\limits_{o \in \mathcal{N}(i)} \exp\left(\frac{1}{\sqrt{d}} {\mathcal{Q}_i^{(l)}}^T \mathcal{K}_o^{(l)}\right)}.
\end{align}
Finally, we use the similarity $\eta_{j,i}^{(l)}$ to calibrate the values $\mathcal{V}_j^{(l)} \in \mathbb{R}^{d}$
\begin{equation}
    \tilde{\mathbf{v}}_j^{(l)} = \eta_{j,i}^{(l)} {\mathcal{V}_j^{(l)}} = \eta_{j,i}^{(l)} (\mathbf{W}_\mathcal{V}^{(l)} \bar{\mathbf{v}}_j^{(l)} + \mathbf{b}_\mathcal{V}^{(l)}),
\end{equation}
where $\tilde{\mathbf{v}}_j^{(l)} \in \mathbb{R}^{d}$ contains attention-aware messages of the neighbor $\mathbf{v}_j^{(l)}$. $\mathbf{W}_\mathcal{V}^{(l)} \in \mathbb{R}^{d \times d_l}$ and $\mathbf{b}_\mathcal{V}^{(l)} \in \mathbb{R}^{d}$ are learnable weights. To implement $\varphi(\cdot)$ in~\Cref{eq:node_rep2}, we aggregate the learned messages from all neighbors of $\mathbf{v}_i^{(l)}$ by simply adding up $\left\{\tilde{\mathbf{v}}_j^{(l)} \mid j \in \mathcal{N}(i)\right\}$. Therefore, the updated node representation $\mathbf{v}_i^{(l+1)} \in \mathbb{R}^{d}$ for the next iteration can be acquired by rewriting $\gamma(\cdot)$ of~\Cref{eq:node_rep3} as 
\begin{equation}
    \mathbf{v}_i^{(l+1)} = \mathbf{W}_2^{(l)} \mathbf{v}_i^{(l)} + \sum_{j \in \mathcal{N}(i)} \tilde{\mathbf{v}}_j^{(l)},
\end{equation}
where $\mathbf{W}_2^{(l)} \in \mathbb{R}^{d \times d_l}$ is a learnable matrix. 
We name the convolution operator discussed above \textbf{TEA}, which stands for Graph \textbf{T}ransformer with \textbf{E}dge \textbf{A}ttributes. As the general pattern of feed-forward networks, we design our network by stacking multiple layers and apply batch normalization~\cite{ioffe2015batch} before nonlinear activations. 

In $G_1$, the updated information directionally flows to certain nodes due to directed connections, such that some nodes carry more informative features while some do not, thus suggesting a strategy to dynamically compact the graph based on the global importance of nodes in the iterative process. With the same purpose as pooling layers in convolutional neural networks (CNNs)~\cite{lecun1998gradient}, graph pooling restructures a coarser graph, where only representative nodes are preserved~\cite{simonovsky2017dynamic,lee2019self,bianchi2022pyramidal}. After acquiring the updated node representation $\mathbf{v}_i^{(l+1)}$, we apply graph pooling to the activated graph representation $\mathbf{v}_G^{(l+1)} = \left\{\mathbf{v}_i^{(l+1)} \mid \mathbf{v}_i \in V\right\} \in \mathbb{R}^{M \times d}$ by incorporating our TEA module denoted by $\theta(\cdot)$ in SAGPool~\cite{lee2019self}. Specifically, we first compute the activated attention mask $\mathbf{z}^{(l+1)} \in \mathbb{R}^{M \times 1}$ via
\begin{equation}
        \mathbf{z}^{(l+1)} = \sigma \bigg(\theta\left(\mathbf{v}_G^{(l+1)}, A^{(l+1)}\right)\bigg),
\end{equation}
where $A^{(l+1)} \in \mathbb{R}^{M \times M \times d_e}$ is the adjacency matrix with edge attributes. Then, based on the top-rank function $f_u(\cdot)$~\cite{knyazev2019understanding}, we are informed of the index $\mathbf{u}$ of top-$u$ nodes ranked in line with the importance by
\begin{equation}
    \mathbf{u} = f_u\left(Z^{(l+1)}\right).
\end{equation}
The top $u$ nodes that contain the most informative features are preserved for the succeeding learning, while the rest nodes are discarded
\begin{align}
        \hat{\mathbf{v}}_G^{(l+1)} &= \left(\mathbf{v}_G^{(l+1)} \odot Z^{(l+1)}\right)_\mathbf{u},\\
        \hat{A}^{(l+1)} &= A^{(l+1)}_{\mathbf{u},\mathbf{u}},
\end{align}
where $\hat{\mathbf{v}}_G^{(l+1)} \in \mathbb{R}^{u \times d}$ and $\hat{A}^{(l+1)} \in \mathbb{R}^{u \times u \times d_e}$ are the updated graph representation and adjacency matrix for the next iteration, respectively. In the final iteration, we exploit global mean pooling to aggregate all nodes for the graph-level output, which is followed by fully connected layers for classification.

\section{Experiments}\label{sec:experiments}
\begin{figure}[t]
    \centering
    \subfloat[Running]{
        \includegraphics[width=\snkth]{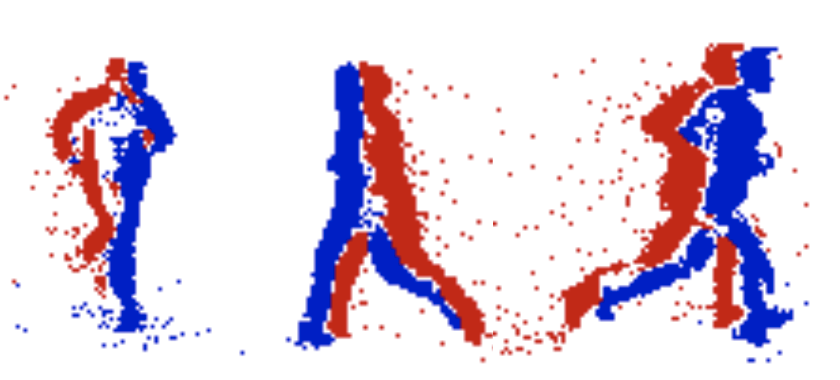}}\hfil
    \subfloat[Walking]{
        \includegraphics[width=\snkth]{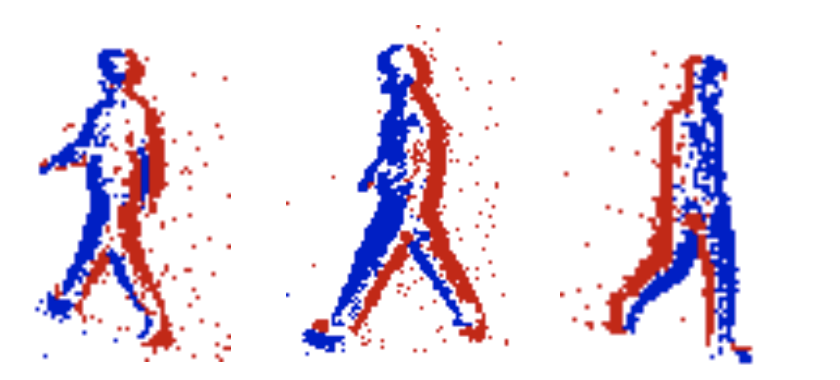}}\hfil
    \subfloat[Jogging]{
        \includegraphics[width=\snkth]{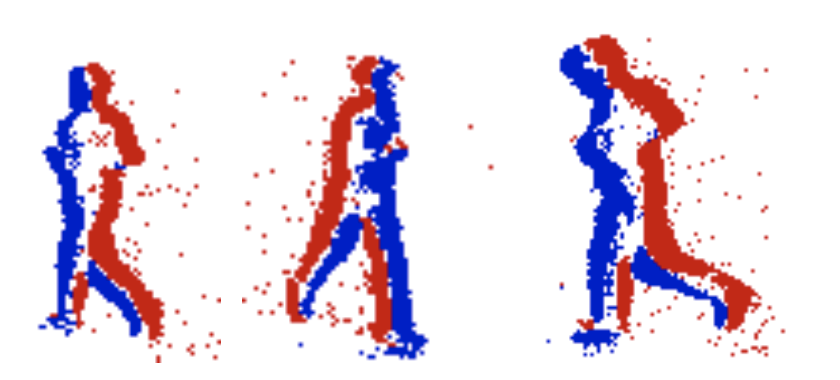}}\hfil
    \subfloat[Hand-clapping]{
        \includegraphics[width=\snkth]{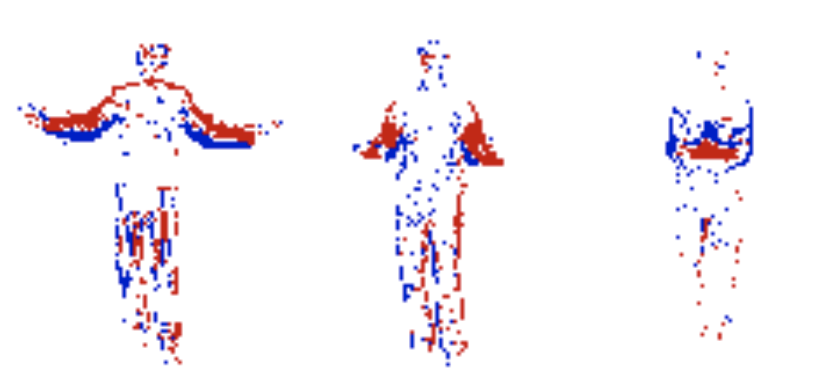}}\hfil
    \subfloat[Hand-waving]{
        \includegraphics[width=\snkth]{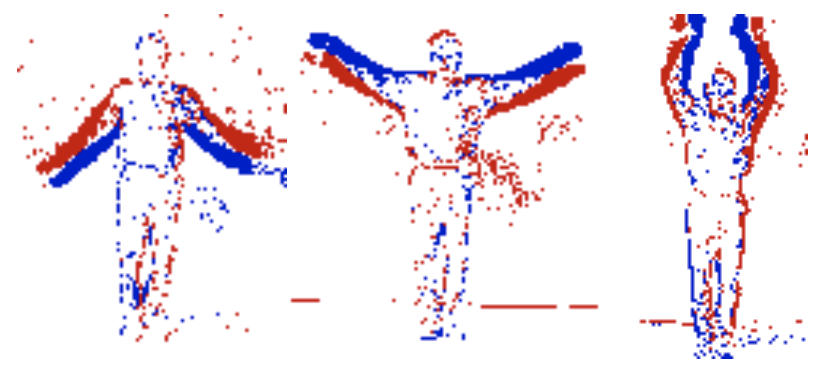}}\hfil
    \subfloat[Boxing]{
        \includegraphics[width=\snkth]{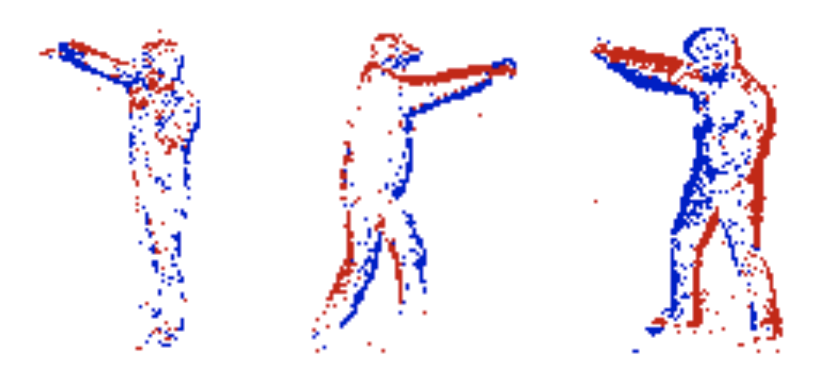}}\hfil
    \caption{Six kinds of human actions in SN-KTH. Red and blue represent events of different polarity. Best viewed in color.}
    \label{fig:snkth}
\end{figure}
\subsection{Neuromorphic Data}
A number of neuromorphic datasets (\eg, N-MNIST~\cite{orchard2015converting} and CIFAR10-DVS~\cite{li2017cifar10}) were recorded and collected by moving either a neuromorphic camera or the images on a monitor, and they thus have clean event data. The N-MNIST dataset is a good benchmark to evaluate the performance of a new classification algorithm. Nevertheless, its events hardly have realistic motion patterns, leading to the models trained on it may not generalize well on other challenging real dynamic scenes. Therefore, it is preferable to use more complex datasets, such as the large dataset ASL-DVS~\cite{bi2019graph} that presents hand-shape recordings under laboratory settings, and N-Cars~\cite{sironi2018hats} that captures real-world driving scenarios.

Several frame-to-event simulators have been developed to facilitate the collection and labelling of event data~\cite{bi2017pix2nvs, hu2021v2e}. To assess the generalizability of our approach to various challenging cases, we simulate a neuromorphic version SN-KTH, built upon the KTH~\cite{schuldt2004recognizing} that presents six kinds of human actions in real scenarios in frame-based format. Specifically, the KTH contains $6$ balanced classes, and each class has $100$ video samples. We split each sample into a number of clips such that each clip only contains a small fraction of the action. Then, we convert these chips to the corresponding event streams. We also remove the clips that only have the background, finally yielding $6$ heavily unbalanced classes in the simulated data. Some samples are shown in~\Cref{fig:snkth}. 

We experiment on real datasets N-Cars~\cite{sironi2018hats}, N-MNIST~\cite{orchard2015converting} and ASL-DVS~\cite{bi2019graph}, and the simulated one SN-KTH. Apart from constructing event representations
on complete event streams, we simulate such a challenging situation by learning a small number of events and evaluating the event streams of varying lengths. For instance, we sample successive $100$ events $\boldsymbol{\epsilon}_{100}$ to construct an event representation for training, and we then test on $\boldsymbol{\epsilon}_{100}$, $\boldsymbol{\epsilon}_{50}$ and $\boldsymbol{\epsilon}_{10}$ ($\boldsymbol{\epsilon}_{50} \subset \boldsymbol{\epsilon}_{100}$ and $\boldsymbol{\epsilon}_{10} \subset \boldsymbol{\epsilon}_{50}$), respectively. As such, each dataset has $1$ training set and $3$ test sets. These settings work under the assumption that the data are correctly collected under normal camera configurations such that not all events in the used short event streams (\ie, $\boldsymbol{\epsilon}_{100}$, $\boldsymbol{\epsilon}_{50}$ and $\boldsymbol{\epsilon}_{10}$) are noise. All experiments are implemented using the PyTorch framework on a machine with an Ubuntu $20.04$ system, a $4.3$ \si{\giga\hertz} AMD Ryzen 3955WX CPU, $4 \times 32$ \si{\giga\byte} RAM, and an Nvidia Tesla V100 GPU with $32$ \si{\giga\byte} memory.
\begin{table*}[t]
\centering
\caption{Comparison on classification performance. \# represents a random guess. The best results are \textbf{highlighted}. The value in parentheses indicates the drop in inference accuracy relative to training accuracy.}
\begin{tabular}{llc|c|c|c}
\toprule
\textbf{Method} & \textbf{Model} & \multicolumn{4}{c}{\textbf{Test}} \\ \midrule
& & N-Cars & N-MNIST & ASL-DVS & SN-KTH \\
\cmidrule(lr){3-3} \cmidrule(lr){4-4} \cmidrule(lr){5-5} \cmidrule(lr){6-6}
 Rebecq~\etal~\cite{rebecq2017real} & ResNet-18 & 0.87~\tiny (0.06) & 0.93~\tiny (0.03) & 0.82~\tiny (0.07) &  0.72~\tiny (0.09) \\ 
 Sironi~\etal~\cite{sironi2018hats} & ResNet-18 & 0.86~\tiny (0.10) & 0.94~\tiny (0.03) & 0.78~\tiny (0.08) & 0.75~\tiny (0.09) \\ 
 Gehrig~\etal~\cite{gehrig2019end} & ResNet-18 & 0.89~\tiny (0.07) & \textbf{0.96}~\tiny (0.03) & 0.84~\tiny (0.03) & 0.76~\tiny (0.06) \\ 
 Wang~\etal~\cite{wang2022exploiting} & ViT & 0.83~\tiny (0.11) & 0.94~\tiny (0.02) & 0.85~\tiny (0.05) & 0.70~\tiny (0.03) \\ 
 Bi~\etal~\cite{bi2019graph} & RG-CNN & 0.88~\tiny (0.06) & 0.94~\tiny (0.03) & 0.88~\tiny (0.05) & 0.74~\tiny (0.06) \\ \midrule
 $G_1$ (ours)& TEA & \textbf{0.91}~\tiny (0.03) & \textbf{0.96}~\tiny (0.01) & \textbf{0.89}~\tiny (0.03) & \textbf{0.80}~\tiny (0.04) \\ \midrule
 \# & \slash & 0.50 & 0.10 & 0.04 & 0.16 \\
\bottomrule
\end{tabular}
\label{tab:cp}
\end{table*}

\begin{table}[t]
\centering
\caption{Comparison on classification with short event streams. The best results are \textbf{highlighted}. The value in parentheses indicates the drop in inference accuracy relative to training accuracy.}
\begin{tabular}{lccc|ccc}
\toprule
\textbf{Method} & \multicolumn{6}{c}{\textbf{Test}} \\ \midrule
& \multicolumn{3}{c}{N-Cars} & \multicolumn{3}{c}{SN-KTH} \\
& $\boldsymbol{\epsilon}_{100}$ & $\boldsymbol{\epsilon}_{50}$ & $\boldsymbol{\epsilon}_{10}$ & $\boldsymbol{\epsilon}_{100}$ & $\boldsymbol{\epsilon}_{50}$ & $\boldsymbol{\epsilon}_{10}$ \\ 
 \cmidrule(lr){2-4} \cmidrule(lr){5-7}
 Rebecq~\etal~\cite{rebecq2017real} & 0.77~\tiny (0.06) & 0.69 & 0.54 & 0.64~\tiny (0.10) & 0.43 & 0.22 \\ 
 Maqueda~\etal~\cite{maqueda2018event} & 0.80~\tiny (0.09) & 0.65 & 0.54 & 0.61~\tiny (0.07) & 0.45 & 0.21 \\ 
 Sironi~\etal~\cite{sironi2018hats} & 0.82~\tiny (0.07) & 0.65 & 0.52 & 0.69~\tiny (0.03) & 0.36 & 0.28 \\ 
 Gehrig~\etal~\cite{gehrig2019end} & 0.85~\tiny (0.06) & 0.72 & 0.54 & 0.70~\tiny (0.01) & 0.40 & 0.26 \\ 
 Bi~\etal~\cite{bi2019graph} & 0.85~\tiny (0.10) & 0.65 & 0.58 & 0.67~\tiny (0.07) & 0.48 & 0.33 \\ \midrule
 $G_1$ (ours)& \textbf{0.86}~\tiny (0.04) & \textbf{0.79} & \textbf{0.67} & \textbf{0.72}~\tiny (0.03) & \textbf{0.57} & \textbf{0.41} \\
\midrule
& \multicolumn{3}{c}{N-MNIST} & \multicolumn{3}{c}{ASL-DVS} \\
& $\boldsymbol{\epsilon}_{100}$ & $\boldsymbol{\epsilon}_{50}$ & $\boldsymbol{\epsilon}_{10}$ & $\boldsymbol{\epsilon}_{100}$ & $\boldsymbol{\epsilon}_{50}$ & $\boldsymbol{\epsilon}_{10}$ \\ 
 \cmidrule(lr){2-4} \cmidrule(lr){5-7}
 Rebecq~\etal~\cite{rebecq2017real} & 0.93~\tiny (0.04) & 0.82 & 0.35 & 0.51~\tiny (0.03) & 0.31 & 0.10 \\ 
 Maqueda~\etal~\cite{maqueda2018event} & 0.93~\tiny (0.04) & \textbf{0.84} & 0.32 & 0.51~\tiny (0.01) & 0.29 & 0.10 \\ 
 Sironi~\etal~\cite{sironi2018hats} & \textbf{0.95}~\tiny (0.03) & 0.81 & 0.33 & \textbf{0.52}~\tiny (0.01) & 0.29 & 0.08 \\ 
 Gehrig~\etal~\cite{gehrig2019end}  & \textbf{0.95}~\tiny (0.02) & 0.82 & 0.29 & \textbf{0.52}~\tiny (0.04) & 0.30 & 0.11 \\ 
 Bi~\etal~\cite{bi2019graph} & 0.92~\tiny (0.01) & 0.80 & 0.33 & 0.50~\tiny (0.07) & 0.32 & 0.11 \\ \midrule
 $G_1$ (ours)& 0.92~\tiny (0.03) & 0.81 & \textbf{0.43} & 0.50~\tiny (0.05) & \textbf{0.40} & \textbf{0.15}  \\
\bottomrule

\end{tabular}
\label{tab:cps}
\end{table}
\begin{table*}[t]
\centering
\caption{Comparison of graph convolution operators. The best results are \textbf{highlighted}. The value in parentheses indicates the drop in inference accuracy relative to training accuracy.}
\begin{tabular}{lcc|c|c|c}
\toprule
\textbf{Operator} & \multicolumn{5}{c}{\textbf{Test}} \\ \midrule
 &  & \multicolumn{1}{c}{N-Cars} & \multicolumn{1}{c}{SN-KTH} & \multicolumn{1}{c}{N-MNIST} & \multicolumn{1}{c}{ASL-DVS} \\
 \cmidrule(lr){3-3} \cmidrule(lr){4-4} \cmidrule(lr){5-5} \cmidrule(lr){6-6}
Simonovsky~\etal~\cite{simonovsky2017dynamic} &  & 0.84~\tiny (0.04) & 0.67~\tiny (0.06) & 0.90~\tiny (0.01) & 0.48~\tiny (0.05) \\ 
Fey~\etal~\cite{fey2018splinecnn} & & 0.83~\tiny (0.05) & 0.70~\tiny (0.09) & 0.88~\tiny (0.03) & 0.48~\tiny (0.07) \\ 
Ranjan~\etal~\cite{ranjan2020asap} & &0.84~\tiny (0.01) & 0.70~\tiny (0.05) & 0.88~\tiny (0.01) & 0.49~\tiny (0.03) \\ \midrule
TEA (without edge attributes) & & 0.83~\tiny (0.04) & 0.65~\tiny (0.04) & 0.88~\tiny (0.01) & 0.48~\tiny (0.03) \\
TEA (ours) & & \textbf{0.86}~\tiny (0.04) & \textbf{0.72}~\tiny (0.03) & \textbf{0.92}~\tiny (0.03) & \textbf{0.50}~\tiny (0.05) \\
\bottomrule
\end{tabular}
\label{tab:glo}
\end{table*}
\subsection{Comparison on Classification Performance}
We compare our approach with several batch-events-based counterparts on classification. A common practice is to perform an event-to-grid conversion that yields grid-based representations as the input of CNNs. We employ ResNet-18~\cite{he2016deep} to test some state-of-the-art event-based grids~\cite{sironi2018hats,rebecq2017real,gehrig2019end}, and use ViT~\cite{dosovitskiy2020vit} to evaluate the recent solution~\cite{wang2022exploiting}. We also compare ours with another graph-based method~\cite{bi2019graph} contributing a network RG-CNN and an event-based radius-neighborhood graph. 

The results in~\Cref{tab:cp} show that our method achieves the best scores for most of the datasets. Especially, we gain at least $8\%$ over \textit{Rebecq's} and \textit{Wang's} on SN-KTH. One of the potential reasons is the loss of temporal details in the operation that transforms events into frames. Temporal information contributes to distinguish similar objects since the pattern of the triggered events of different objects is varying. For example, \textit{running} and \textit{jogging} look similar in static frames (as \Cref{fig:snkth}), which tends to confuse the classifier to make incorrect estimations. However, \textit{running} typically has a shorter temporal distance between two events since it moves faster. Similar cases also occur in other used datasets. Our simple yet effective scheme allows nodes to be connected in chronological order, as they are present in the original event stream, and the nodes are associated by the edges featured by the temporal distance, leveraging the temporal information of asynchronous events in the synchronous graph. The proposed operator TEA is also temporal-wise (\ie, \Cref{eq:pre-cali}) and is thus robust to such challenging cases.

\subsection{Classification with Short Event Streams}
The negatives of batch-events-based methods are the delays caused and the computational resources consumed when gathering a large batch of events over a long time period. However, some applications (\eg, those for mobile devices and self-driving cars) expect the minimum runtime and resources consumption, which requires a neuromorphic system to perform accurate classification on as few events as possible. As described above, we train on successive $100$ events $\boldsymbol{\epsilon}_{100}$ that can be collected within a short period by a neuromorphic camera, and then we test on much fewer events. 

\Cref{tab:cps} presents the accuracy comparison when short event streams are used. We employ $\boldsymbol{\epsilon}_{100}$ in the training set to shape and train different event representations, and use test samples $\boldsymbol{\epsilon}_{100}$, $\boldsymbol{\epsilon}_{50}$ and $\boldsymbol{\epsilon}_{10}$ for inference, while we have $\boldsymbol{\epsilon}_{50} \subset \boldsymbol{\epsilon}_{100}$ and $\boldsymbol{\epsilon}_{10} \subset \boldsymbol{\epsilon}_{50}$. We expect that the system can give correct estimations based on a small fragment of information after observing global contents. In the comparisons, our solution achieves better results in most of the test cases. It is noticeable that, when testing on $\boldsymbol{\epsilon}_{50}$ and $\boldsymbol{\epsilon}_{10}$, most competing approaches suffer from a considerable drop in accuracy due to less information available in a batch. For example, we outperform \textit{Sironi's} on $\boldsymbol{\epsilon}_{50}$ of N-Cars and \textit{Gehrig's} on $\boldsymbol{\epsilon}_{10}$ of N-MNIST by $14\%$, showing that we are better at the demanding situation in which data are limited, and suffer much less from the loss of information. Compared with the frame-based representations where the loss of events may greatly corrupt the contents, graphs are shown to be more robust especially when the temporal-wise connectivity is employed, since the order in which information is expressed is hardly disturbed. This study demonstrates that, when using very few events for the reduction of the required runtime and resources consumption, our method is still capable of performing accurate classification.

\subsection{Comparison of Graph Learning Operators}
We compare several competing methods in experiments to verify the rationality and superiority of the proposed graph operator. We simply replace TEA with other operators and train the networks to learn $G_1$ for the same epochs. Moreover, we investigate the impact of edge attributes in the self-attention mechanism, that is $\bar{\mathbf{v}}_j^{(l)} = \mathbf{v}_j^{(l)}$ in~\Cref{eq:pre-cali}. \Cref{tab:glo} shows that ours outstrips other operators in all the datasets used, which demonstrates the effectiveness of the dot-product attention in event-based graph learning. Using TEA to iteratively pass information from past events to current events is consistent with the moving pattern of objects where events are triggered sequentially in space-time. An issue of the proposed strategies to build graphs is that adjacent events connected in the graph by hard-coded rules do not necessarily have any underlying relationship. Before computing the similarity, we thus learn dynamic edge weights from static edge attributes to calibrate neighbors for obtaining more discriminative node representations, as we do in~\Cref{eq:pre-cali}, which acts as adaptive calibration of the hard-coded connections when cooperating with graph pooling. When the self-attention does not involve edge attributes, the drop in accuracy reflects the importance of edge information and the necessity of dynamic calibration for node representations and hard-coded connections.
\begin{table}[t]
\centering
\caption{Comparison on the memory footprint (in Megabyte).}
\begin{tabular}{lccc}
\toprule
 \textbf{Data} & \multicolumn{3}{c}{\textbf{Representation}} \\ \midrule
 & Rebecq~\etal~\cite{rebecq2017real} & Bi~\etal~\cite{bi2019graph} & $G_1$ (ours) \\ \cmidrule(lr){2-2} \cmidrule(lr){3-3} \cmidrule(lr){4-4}
N-Cars & 86.4 & 67.3 & 25.7 \\
SN-KTH & 248.8 & 55.1 & 25.0 \\
N-MNIST & 16.9 & 48.5 & 28.9 \\
ASL-DVS & 259.2 & 78.1 & 23.8 \\
\bottomrule
\end{tabular}
\label{tab:mf}
\end{table}

\subsection{Comparison on the Memory Footprint}
Mobile facilities are typically memory constrained. We compare the data size of some training samples ($\boldsymbol{\epsilon}_{100}$) of different event representations in each dataset (as shown in \Cref{tab:mf}). Except for N-MNIST, the data size of graphs is much smaller than that of grid-based representations (\ie, single-channel event frames). The reason is that the grids of N-MNIST have $28 \times 28$ pixels, while those of SN-KTH and ASL-DVS have over $150 \times 150$ pixels. However, only $100$ pixels associated with $\boldsymbol{\epsilon}_{100}$ are informative, while the rest without information still takes up space. Besides, not all frames converted from the same number of events in a particular dataset have the same dimensions, and more processing has to be done. These issues exist in other grid-based representations that exploit the spatial position of events~\cite{sironi2018hats,maqueda2018event,gehrig2019end,kim2021n}. In contrast, our $G_1$ saves more on the memory footprint and has an efficient event transformation invariant to the camera resolution, which is more favorable to mobile applications.

\subsection{Comparison on Runtime}
In~\Cref{tab:rt}, we compare the average runtime of different event representations shaped by varying numbers of events. The runtime of a classification system involves the duration of event collection by a camera and the delay of event transformation. It can be observed that the number of events necessary for the system greatly influences the latency (\ie, $5$ \si{\ms} of $\boldsymbol{\epsilon}_{all}$ \vs $2$ \si{\us} of $\boldsymbol{\epsilon}_{10}$), and hence it is preferable to having good classification performance on short event streams. Mapping events to the target representation is a decisive factor for time cost, especially when a huge number of events are involved. Compared with the competing methods, particularly \textit{Bi's} whose graphs typically have redundant connections due to the radius-based connectivity, ours necessitates lower delays in event transformation.
\begin{table*}[t]
\centering
\caption{Comparison on runtime (duration of event collection + delay of event transformation). $\boldsymbol{\epsilon}_{all}$ indicates all the events of a sample.}
\begin{tabular}{lc|c|c}
\toprule
\textbf{Method} & \multicolumn{3}{c}{\textbf{Runtime}} \\ \midrule
& \multicolumn{1}{c}{$\boldsymbol{\epsilon}_{all}$} & \multicolumn{1}{c}{$\boldsymbol{\epsilon}_{100}$} & \multicolumn{1}{c}{$\boldsymbol{\epsilon}_{10}$}\\
 \cmidrule(lr){2-2} \cmidrule(lr){3-3} \cmidrule(lr){4-4}
 Sironi~\etal~\cite{sironi2018hats} & 5 \si{\ms} + 1.5 \si{\s} & 18 \si{\us} + 2.9 \si{\ms} & 2 \si{\us} + 0.8 \si{\ms} \\
 Bi~\etal~\cite{bi2019graph} & 5 \si{\ms} + 2.2 \si{\s} & 18 \si{\us} + 5.1 \si{\ms} & 2 \si{\us} + 1.3 \si{\ms} \\ 
 $G_1$ (ours) & 5 \si{\ms} + 0.8 \si{\s} & 18 \si{\us} + 1.9 \si{\ms} & 2 \si{\us} + 0.3 \si{\ms} \\
\bottomrule
\end{tabular}
\label{tab:rt}
\end{table*}
\begin{table*}[t]
\centering
\caption{Top-1 classification accuracy. \# represents a random guess. The best results are \textbf{highlighted}.}
\begin{tabular}{cccc|ccc|ccc|ccc}
\toprule
\textbf{Graph} & \multicolumn{12}{c}{\textbf{Test}} \\ \midrule
& \multicolumn{3}{c}{N-Cars} & \multicolumn{3}{c}{SN-KTH} & \multicolumn{3}{c}{N-MNIST} & \multicolumn{3}{c}{ASL-DVS} \\
& $\boldsymbol{\epsilon}_{100}$ & $\boldsymbol{\epsilon}_{50}$ & $\boldsymbol{\epsilon}_{10}$ & $\boldsymbol{\epsilon}_{100}$ & $\boldsymbol{\epsilon}_{50}$ & $\boldsymbol{\epsilon}_{10}$ & $\boldsymbol{\epsilon}_{100}$ & $\boldsymbol{\epsilon}_{50}$ & $\boldsymbol{\epsilon}_{10}$ & $\boldsymbol{\epsilon}_{100}$ & $\boldsymbol{\epsilon}_{50}$ & $\boldsymbol{\epsilon}_{10}$ \\ 
 \cmidrule(lr){2-4} \cmidrule(lr){5-7} \cmidrule(lr){8-10} \cmidrule(lr){11-13}
 $G_1$ & \textbf{0.83} & \textbf{0.77} & \textbf{0.66} & \textbf{0.67} & \textbf{0.53} & 0.40 & \textbf{0.88} & \textbf{0.77} & \textbf{0.42} & \textbf{0.48} & 0.35 & \textbf{0.15} \\ 
 $G_2$ & 0.82 & 0.71 & 0.58 & 0.66 & 0.50 & \textbf{0.41} & 0.85 & 0.65 & 0.22 & 0.43 & 0.30 & \textbf{0.15} \\ 
 $G_3$ & 0.82 & 0.76 & 0.64 & \textbf{0.67} & \textbf{0.53} & 0.40 & 0.86 & \textbf{0.77} & 0.28 & 0.43 & \textbf{0.36} & 0.14 \\ 
 $G_4$ & 0.63 & 0.52 & 0.50 & 0.36 & 0.27 & 0.17 & 0.39 & 0.21 & 0.13 & 0.26 & 0.17 & 0.06\\\midrule
 \# & 0.50 & 0.50 & 0.50 & 0.16 & 0.16 & 0.16 & 0.10 & 0.10 & 0.10 & 0.04 & 0.04 & 0.04\\
\bottomrule
\end{tabular}
\label{tab:tps}
\end{table*}
\begin{figure*}[t]
    \centering
    \subfloat[N-Cars]{
        \includegraphics[width=\tps]{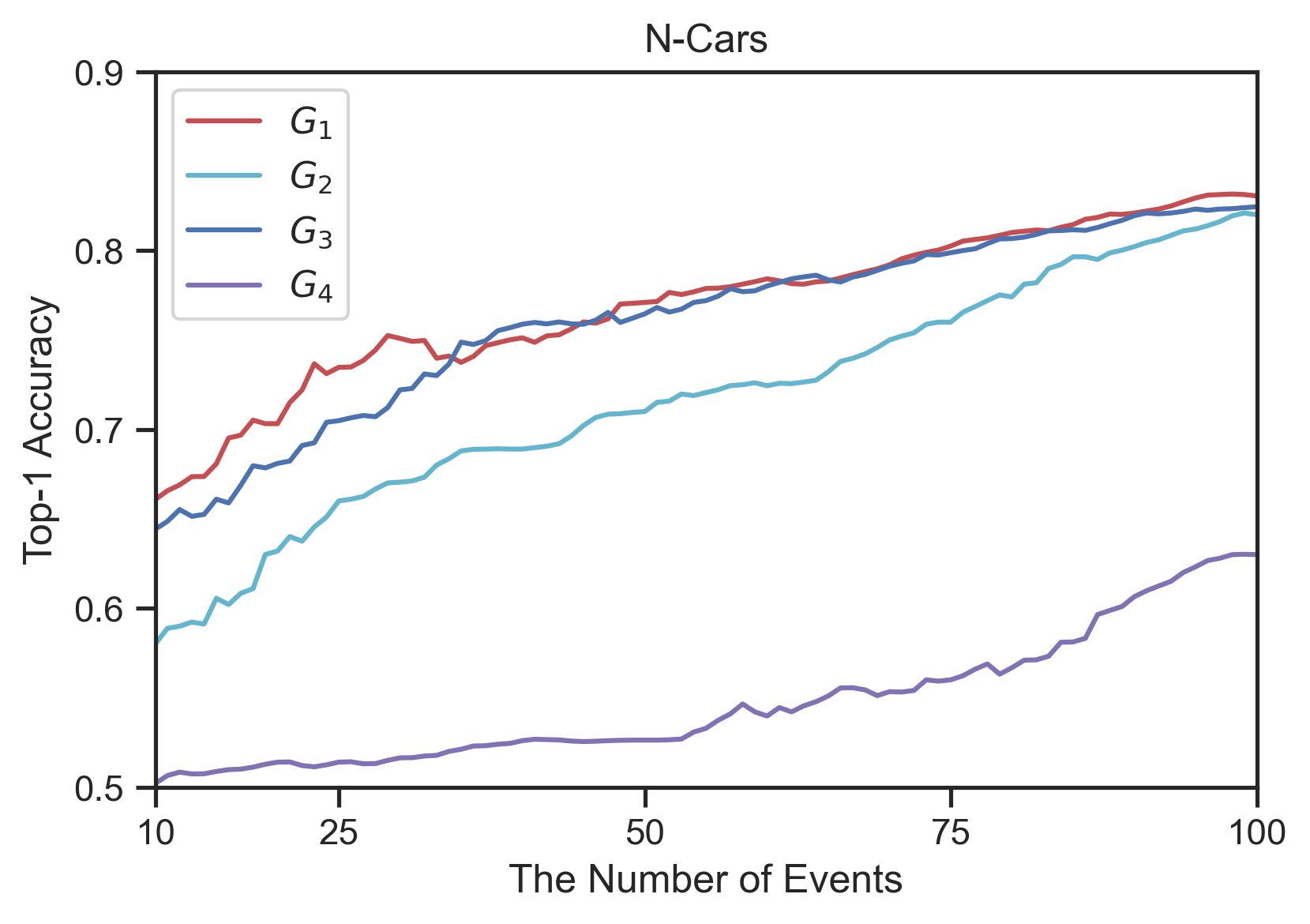}}
    \subfloat[SN-KTH]{
        \includegraphics[width=\tps]{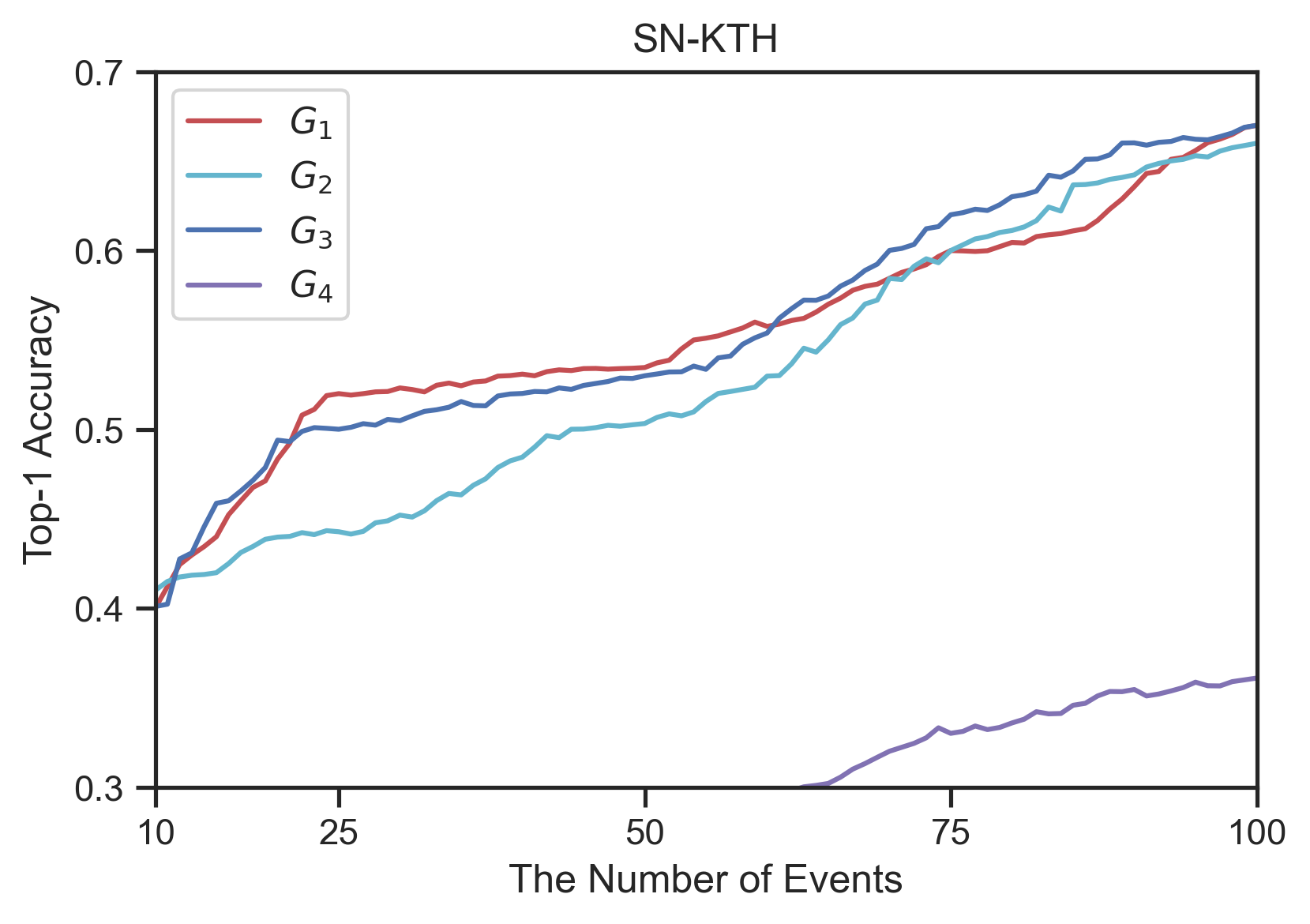}}\\
    \subfloat[N-MNIST]{
        \includegraphics[width=\tps]{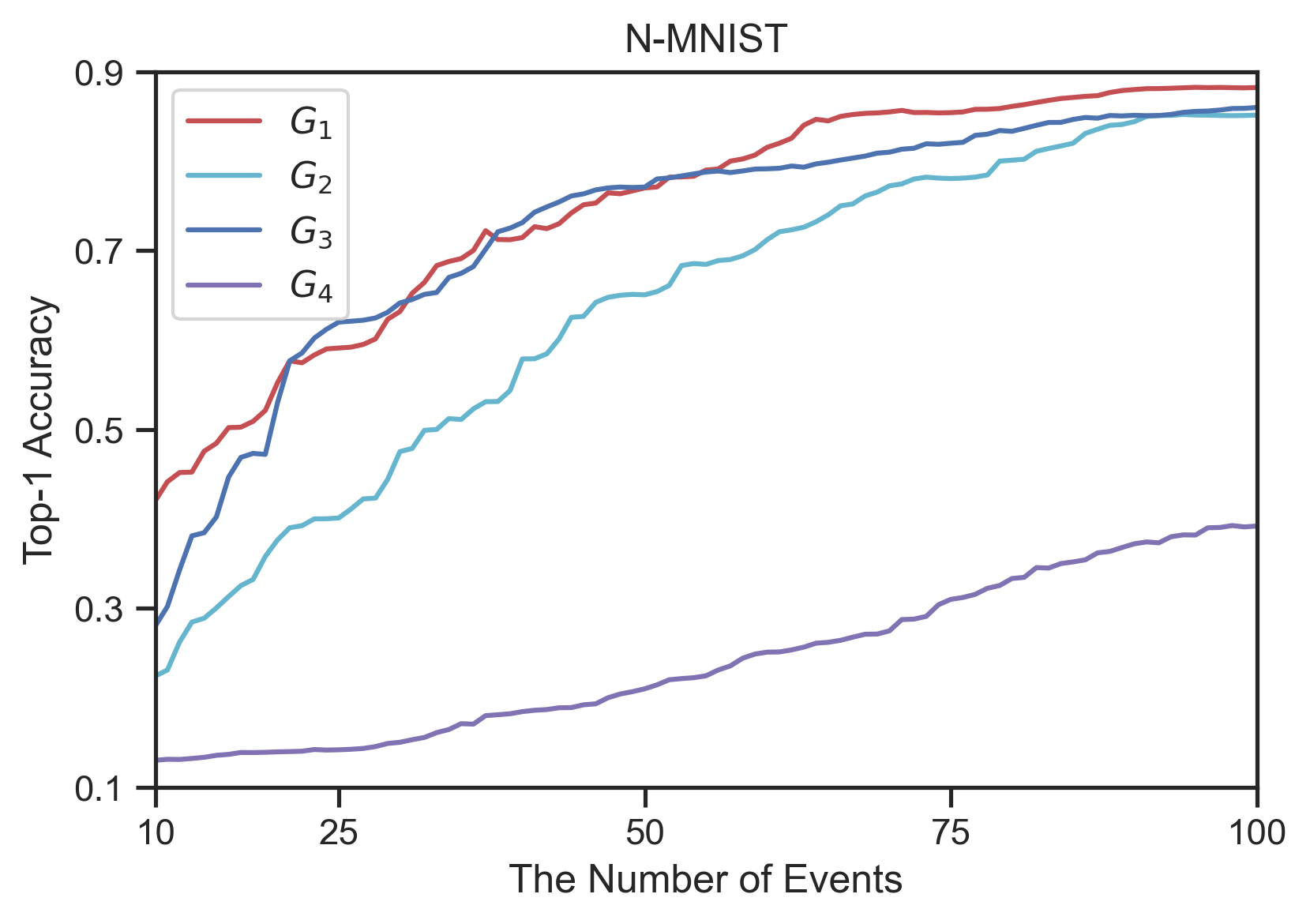}}
    \subfloat[ASL-DVS]{
        \includegraphics[width=\tps]{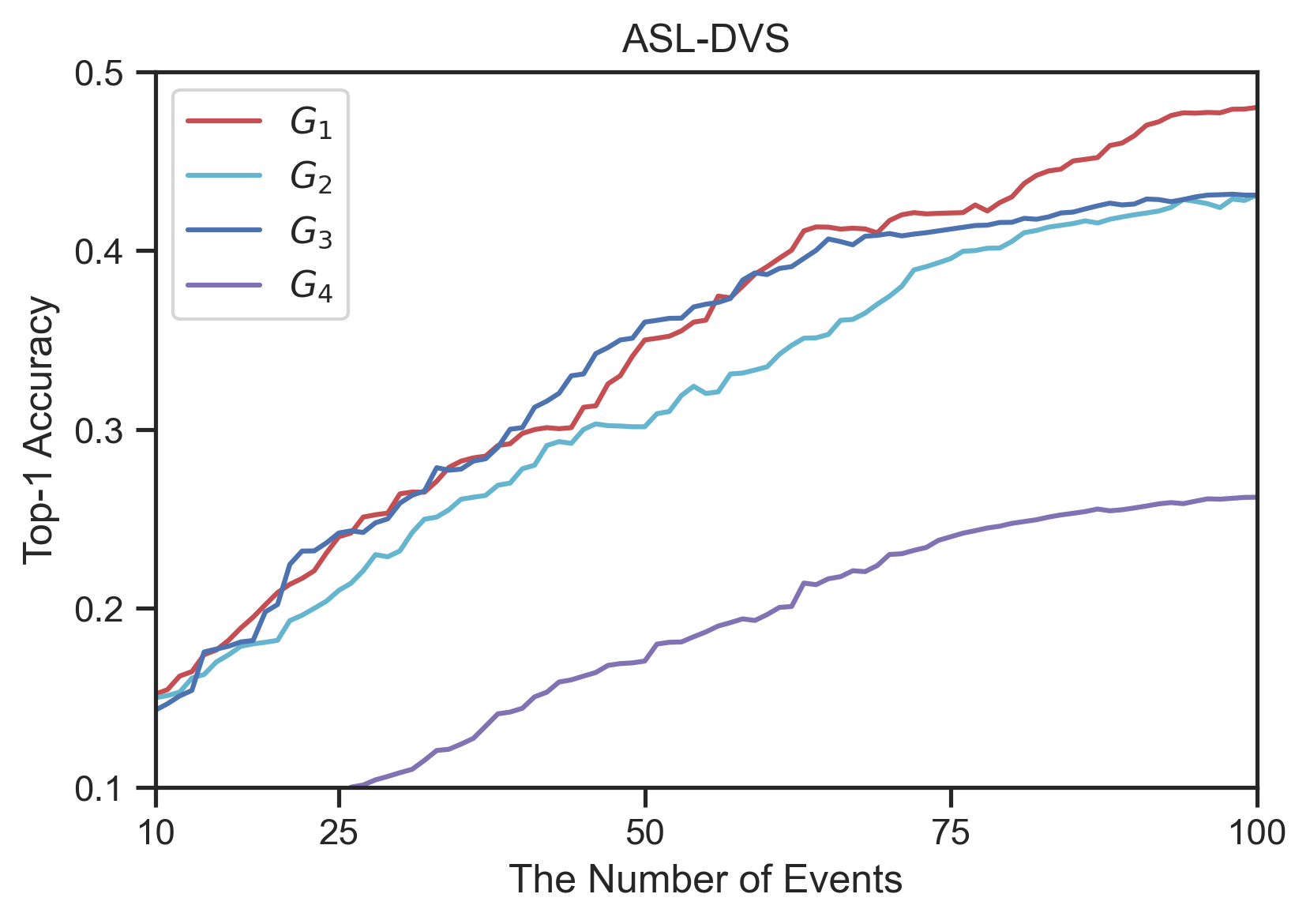}}
        
    \caption{Top-1 classification accuracy as the number of inference events in a batch increases from $10$ to $100$. Best viewed in color and zoom-in.}
    \label{fig:tps}
\end{figure*}

\subsection{Investigations into Temporal, Polar and Spatial Influence}
We propose four strategies to convert events to a graph, involving the directionality of connections and the presence of the properties of events on the graph. \Cref{tab:tps} exhibits that $G_1$ achieves the best results in most of the test cases. Nevertheless, $G_3$ can also obtain mostly comparable results, proving that polarity information does not make graphs significantly more informative and enhance the model discriminability. In contrast, $G_4$ shows a substantial decline in performance due to the lack of the explicit position of events to characterize nodes, such that the models can barely recognize objects from $\boldsymbol{\epsilon}_{10}$, illustrating the significance of the spatial position of nodes for graph-level classification. $G_2$ also fails to be competitive, which suggests that undirected connections in our graphs cannot achieve efficient message passing and thus lead to inefficient learning. More comprehensive comparisons are shown in~\Cref{fig:tps}.

\subsection{Investigations into Temporal Degree}
We use a hyper-parameter $\tau$ to represent temporal degree to enrich the routes for messages passing between the events sharing the same position. As shown in~\Cref{tab:td}, we test $\tau = \{1, 2, 4\}$ and observe that the models achieve higher accuracy in most of the test cases when taking a larger $\tau$. However, some results of $\tau = 4$ are not better than those of $\tau = 2$, while the comparison between $\tau = 2$ and $\tau = 1$ presents more obvious improvements. Besides, $\tau$ has a relatively small effect on $\boldsymbol{\epsilon}_{10}$. The reason could be that fewer events triggered at the same position exist in shorter event streams such that the information conveyed by the graph does not change as $\tau$ varies. The experiment also reflects that events that are temporally distant but not spatially different have underlying relationships even though they do not occur adjacently. 

Furthermore, enlarging $\tau$ can mitigate the impact of noise events that are probably triggered due to current leaks. We quantify the relationship between $\tau$ and noise by injecting uniformly-distributed $10\%$ noise into test samples that are then transformed into the graphs with different levels of temporal degree. In~\Cref{tab:td}, we use a value in parentheses to represent the drop in inference accuracy on the noisy test samples, relative to the original test samples. The comparisons show that leveraging a large value of $\tau$ can suffer less from the decrease in accuracy.
\begin{table}[t]
\centering
\caption{Top-1 classification accuracy of $G_1$. The best results achieved by the minimum $\tau$ are \textbf{highlighted}. The value in parentheses indicates the drop in inference accuracy on the noisy test samples, relative to the original test samples.}
\begin{tabular}{cccc|ccc}
\toprule
\textbf{Temporal Degree} & \multicolumn{6}{c}{\textbf{Test}} \\ \midrule
& \multicolumn{3}{c}{N-Cars} & \multicolumn{3}{c}{SN-KTH} \\
& $\boldsymbol{\epsilon}_{100}$ & $\boldsymbol{\epsilon}_{50}$ & $\boldsymbol{\epsilon}_{10}$ & $\boldsymbol{\epsilon}_{100}$ & $\boldsymbol{\epsilon}_{50}$ & $\boldsymbol{\epsilon}_{10}$ \\ 
 \cmidrule(lr){2-4} \cmidrule(lr){5-7} 
 $\tau = 1$ & 0.83~\tiny (0.04) & 0.77~\tiny (0.02) & 0.66~\tiny (0.01) & 0.67~\tiny (0.05) & 0.53~\tiny (0.02) & 0.40~\tiny (0.00) \\
 $\tau = 2$ & 0.85~\tiny (0.02) & \textbf{0.79}~\tiny (0.02) & 0.66~\tiny (0.00) & 0.68~\tiny (0.05) & 0.55~\tiny (0.01) & \textbf{0.41}~\tiny (0.01) \\ 
 $\tau = 4$ & \textbf{0.86}~\tiny (0.02) & 0.79~\tiny (0.01) & \textbf{0.67}~\tiny (0.00) & \textbf{0.72}~\tiny (0.03) & \textbf{0.57}~\tiny (0.01) & 0.41~\tiny (0.00) \\ 
\midrule
& \multicolumn{3}{c}{N-MNIST} & \multicolumn{3}{c}{ASL-DVS} \\
& $\boldsymbol{\epsilon}_{100}$ & $\boldsymbol{\epsilon}_{50}$ & $\boldsymbol{\epsilon}_{10}$ & $\boldsymbol{\epsilon}_{100}$ & $\boldsymbol{\epsilon}_{50}$ & $\boldsymbol{\epsilon}_{10}$ \\ 
 \cmidrule(lr){2-4} \cmidrule(lr){5-7}
 $\tau = 1$ & 0.88~\tiny (0.01) & 0.77~\tiny (0.02) & 0.42~\tiny (0.01) & 0.48~\tiny (0.03) & 0.35~\tiny (0.01) & 0.15~\tiny (0.00) \\
 $\tau = 2$ & 0.89~\tiny (0.00) & 0.80~\tiny (0.02) & 0.42~\tiny (0.01) & \textbf{0.50}~\tiny (0.03) & \textbf{0.40}~\tiny (0.02) & \textbf{0.16}~\tiny (0.00) \\ 
 $\tau = 4$ & \textbf{0.92}~\tiny (0.00) & \textbf{0.81}~\tiny (0.02) & \textbf{0.43}~\tiny (0.01) & 0.50~\tiny (0.02) & 0.40~\tiny (0.02) & 0.15~\tiny (0.00) \\ 
\bottomrule
\end{tabular}
\label{tab:td}
\end{table}
\subsection{Discussion on Limitations}
We discuss the limitations of the proposed approach in this section. Classifying neuromorphic objects based on short event streams builds upon the assumption that the events used are not all noise. However, neuromorphic cameras are highly sensitive to various interference and can return a lot of noise accordingly. The proposed approach is not naturally robust to noise. Therefore, the quality of the sampled events tends to have a significant impact on accuracy. There are some possible solutions to alleviate such an impact. For example, conducting proper event denoising before sampling, or evaluating the events distributed over dense (\ie, highly dynamic) regions. Furthermore, the dot-product attention involves more operations and higher computation complexity compared to CNNs, leading to more effort in training periods.

\section{Conclusion}\label{sec:conclusion}
In this work, we introduce a simple yet effective graph representation for asynchronous events and a Graph Transformer to perform end-to-end neuromorphic classification. Our solution can achieve higher accuracy in both real and simulated data and perform robustly when events and computational resources (\eg, memory and time) are constrained, indicating that our method is more conducive to the applications embedded in mobile devices. In the future, we will extend our work to further investigate the feasibility of real-time functions. 

\section*{CRediT Author Statement}
\textbf{Pei Zhang:} Conceptualization, Methodology, Software, Formal analysis, Investigation, Data Curation, Writing - Original Draft. \textbf{Chutian Wang:} Writing - Original Draft. \textbf{Edmund Y. Lam:} Conceptualization, Project Administration, Writing - Review \& Editing, Supervision, Funding Acquisition.

\section*{Declaration of Competing Interest}
The authors declare that they have no known competing financial interests or personal relationships that could have appeared to influence the work reported in this paper.

\section*{Acknowledgments}
This work is supported in part by the Research Grants Council of Hong Kong SAR (GRF 17201620, 17200321) and by ACCESS --- AI Chip Center for Emerging Smart Systems, sponsored by InnoHK funding, Hong Kong SAR.




\bibliographystyle{elsarticle-num} 
\bibliography{refs.bib}

\end{document}